\documentclass[conference, a4paper]{IEEEtran}
\IEEEoverridecommandlockouts
\usepackage{cite}
\usepackage{amsmath,amssymb,amsfonts}
\usepackage{algorithmic}
\usepackage{graphicx}
\usepackage{textcomp}
\usepackage{xcolor}
\usepackage{colortbl}
\usepackage{makecell}
\usepackage{adjustbox}
\usepackage{multirow,booktabs}
\usepackage{wrapfig,lipsum,booktabs}
\usepackage[flushleft]{threeparttable}
\usepackage{kotex}
\usepackage{array}
\usepackage{footnote}
\usepackage[misc]{ifsym}
\usepackage{pifont}

\newcommand{\cmark}{\ding{51}}%
\newcommand{\xmark}{\ding{55}}%
\newcommand{\Frst}[1]{\textcolor{red}{\textbf{#1}}}
\newcommand{\Scnd}[1]{\textcolor{blue}{\textbf{#1}}}

\def\BibTeX{{\rm B\kern-.05em{\sc i\kern-.025em b}\kern-.08em
    T\kern-.1667em\lower.7ex\hbox{E}\kern-.125emX}}

\begin{document}

\title{A Hybrid Vision Transformer Approach for Mathematical Expression Recognition}

\author{\IEEEauthorblockN{
Anh Duy Le\IEEEauthorrefmark{1},
Van Linh Pham\IEEEauthorrefmark{1}, 
Vinh Loi Ly\IEEEauthorrefmark{1},
Nam Quan Nguyen\IEEEauthorrefmark{1},
Huu Thang Nguyen\IEEEauthorrefmark{1},
Tuan Anh Tran\IEEEauthorrefmark{2}\IEEEauthorrefmark{3}(\Letter)
}
 \IEEEauthorblockA{\IEEEauthorrefmark{1} \small Viettel Cyberspace Center, Viettel Group, Vietnam.
 \\Lot D26 Cau Giay New Urban Area, Yen Hoa Ward, Cau Giay District, Hanoi, Vietnam.}
\IEEEauthorblockA{\IEEEauthorrefmark{2} \small Faculty of Computer Science $\&$ Engineering, Ho Chi Minh City-University of Technology (HCMUT), \\ 268 Ly Thuong Kiet Street, District 10, Ho Chi Minh City, Vietnam.}
\IEEEauthorblockA{\IEEEauthorrefmark{3} \small Vietnam National University Ho Chi Minh City, Linh Trung Ward, Thu Duc District, Ho Chi Minh City, Vietnam.}
\{leanhduy497, phamvanlinh143, vinhloiit1327, ngnamquan\}@gmail.com, nhthang99@outlook.com, trtanh@hcmut.edu.vn
}

\maketitle

\IEEEpubidadjcol

\begin{abstract}
    One of the crucial challenges taken in document analysis is mathematical expression recognition. Unlike text recognition which only focuses on one-dimensional structure images, mathematical expression recognition is a much more complicated problem because of its two-dimensional structure and different symbol size. In this paper, we propose using a Hybrid Vision Transformer (HVT) with 2D positional encoding as the encoder to extract the complex relationship between symbols from the image. A coverage attention decoder is used to better track attention's history to handle the under-parsing and over-parsing problems. We also showed the benefit of using the [CLS] token of ViT as the initial embedding of the decoder. Experiments performed on the IM2LATEX-100K dataset have shown the effectiveness of our method by achieving a BLEU score of 89.94 and outperforming current state-of-the-art methods. 
\end{abstract}


\section{Introduction}\label{sec:intro}
Mathematical expression recognition is one of the important processes in scientific documents analysis \cite{anh2017}. Despite the importance of this task, solving mathematical expression recognition is still very challenging. One of the reasons for the difficulty of math recognition compared to normal text recognition is that math formula usually has 2-D spatial structure relationship \cite{Yan2021} instead of 1-D ones from normal text data. The spatial structure relationship of math formula is presented by many math symbols such as superscript, subscript, fraction symbol, etc. The traditional approach usually solves this problem in two stages. First, the character segmentation stage is used to segment each character in math formula and then classify it based on the given vocabulary. Second, the structural analysis stage is used to identify the spatial relationships between all characters of the math formula.

Due to the success of sequence to sequence (Seq2seq) architecture \cite{sutskever2014sequence} from machine translation problems, many recent works have applied this architecture to many other fields, including speech recognition \cite{chan2015listen}, text recognition \cite{shi2016end}, image captioning \cite{xu2015show}. Seq2seq architecture includes two main parts: encoder and decoder. Mathematical expression recognition can also be considered a sequence translation problem, where the input, in this case, is the image of math expression and the output is a 1-D sequence of LaTeX. Therefore, we can use the Seq2seq approach to solve the mathematical expression recognition problem. Indeed, recent works have proposed many variants of Seq2seq architecture \cite{Yan2021, deng2016you, zhang2018multi, zhang2019improved, pang2021global} and achieved many promising results. Despite that, the design of these architectures still has many limitations. For example, Deng et al. \cite{deng2016you} introduced a multi-row encoder to capture the non-left-to-right relationships of math symbols better, or Zhang et al. \cite{zhang2018multi} with a multi-scale encoder using DenseNet \cite{huang2017densely} to handle the different size of symbols. Recently, Zelun Wang and Jyh Charn Liu \cite{wang2021translating} had designed a convolutional neural network (CNN) backbone with an additional 2D positional encoding and performed sequence-level learning based on reinforcement learning. These models entirely depend on the feature extracted by a CNN. They so lack global information, which is necessary for modeling spatial relationships between different math symbols since math expressions can contain related symbols which are far apart, limiting them in recognizing long expressions.
\begin{figure}[t!]
    \centering
    \includegraphics[width=0.47\textwidth]{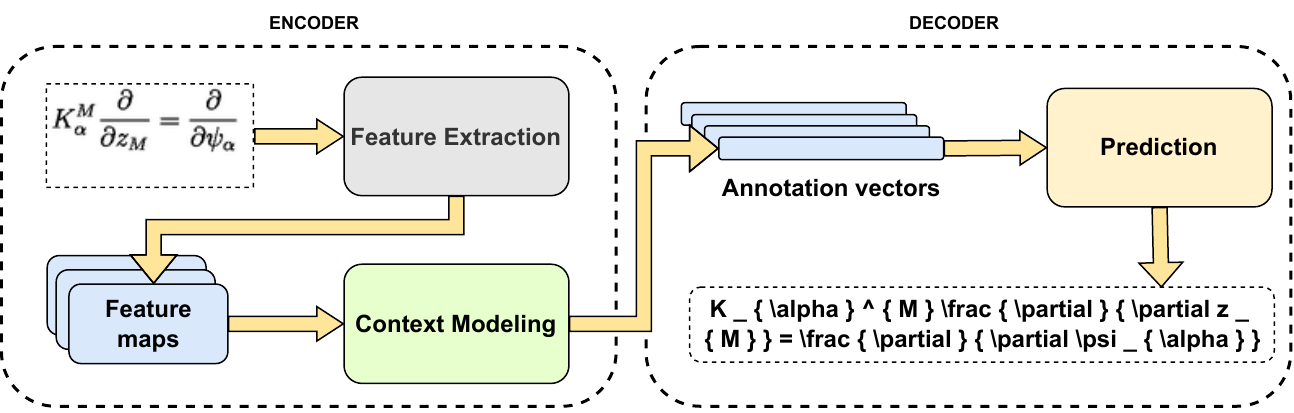}
    \caption{A general overview of our proposed framework} \label{fig:general}
\end{figure}
Inspired by the success of Vision Transformer (ViT) \cite{dosovitskiy2020image} architecture, we propose a novel Hybrid Vision Transformer (HVT) approach to acting as an encoder of the Seq2seq model to alleviate the lacking of global information problem. An HVT consists of a CNN backbone, a 2D positional encoding (2DPE), and a stacking of multiple ViT blocks. Image's features are first extracted by a CNN backbone to reduce input size and get high-level information then are encoded with global information using ViT to return the annotation vectors \cite{zhang2017watch}. 2D positional encoding helps the feature maps reserve more spatial information for both vertical and horizontal directions. For the decoder, we follow the coverage attention idea from \cite{zhang2017watch} by using an additional coverage vector to align the attention weights. Furthermore, we also leverage the [CLS] token embedding of ViT as the initial hidden state for our decoder. In general, our architecture includes three main stages as shown in Fig. \ref{fig:general}: Feature extraction and context modeling using HVT in the encoder and prediction in the decoder. Experiments on benchmark dataset IM2LATEX-100K have shown a competitive result and achieved a new state-of-the-art (SOTA) result with a BLEU score of 89.94, an image exact match rate of 86.48. Our main contributions can be summarized as follows:\\
\indent A novel Hybrid Vision Transformer approach for the encoder of the Seq2seq model.\\
\indent Re-design the Seq2seq framework in both the encoder and the decoder to better suit the math recognition problems and achieve the SOTA result on the IM2LATEX-100K dataset.\\
\indent Extensive ablation experiments and analysis.\\
Our remaining sections are organized as follows: Review the related work, present our proposed method, perform experiments and analyze results, and give conclusions and future work.
\section{Related work}\label{sec:related}
Mathematical expression recognition has been an interesting research topic for a long time. Before the rise of the deep learning era, researchers usually proposed new methods based on two main approaches: rule-based and grammar-based methods. These methods require knowledge of mathematical grammar and tedious work to design suitable rules. From the theory point of view, a mathematical expression recognition problem can be considered an image-based sequence prediction problem that can be solved by a Seq2seq model. With the rise of deep learning, many Seq2seq models \cite{sutskever2014sequence} which learn directly from data are being proposed and achieved better performance than previous non-deep methods. Deng et al. \cite{deng2016you} is considered the first paper to use the Seq2seq model for this problem. In their work, they proposed to use a multi-row encoder to learn the spatial structure of math formulas better. As an improvement from Deng et al., Zhang et al. \cite{zhang2017watch} integrate a coverage vector into the attention module to deal with over parsing in mathematical expression recognition. The coverage vector gives attention module information about the history of alignment in the past to push attention weight to appropriate local regions of feature maps. Zhang et al. \cite{zhang2018multi} have introduced a novel multi-scale encoder to better capture different sizes of math symbols, usually in handwriting math expressions. Using a multi-scale encoder, the model can learn the representation of large and small symbols. Bender et al. \cite{bender2019learning} focused on extracting fine-grained features from math images. In order to handle the neglect of key features when perform attention causing the decoder to give wrong prediction, Li et al. \cite{li2020improving} used a drop attention module to randomly suppress features in training phase, thus, make the model more robust. Yan et al. \cite{Yan2021} proposed a decoder that used a CNN instead of the recurrent network to speed up the training and predicting process. Pang et al. \cite{pang2021global} use a global-context network to aggregate different global features using a global context module integrated into a CNN backbone. Their backbone is not optimized for text-based recognition compared to ours. It has a more well-designed CNN backbone for math recognition and is entirely based on ViT to capture global dependencies and position information.\\

\section{Methodology} \label{sec:proposed}
\subsection{Problem definition}
\begin{figure*}[thbp]
    \includegraphics[width=0.6\textwidth]{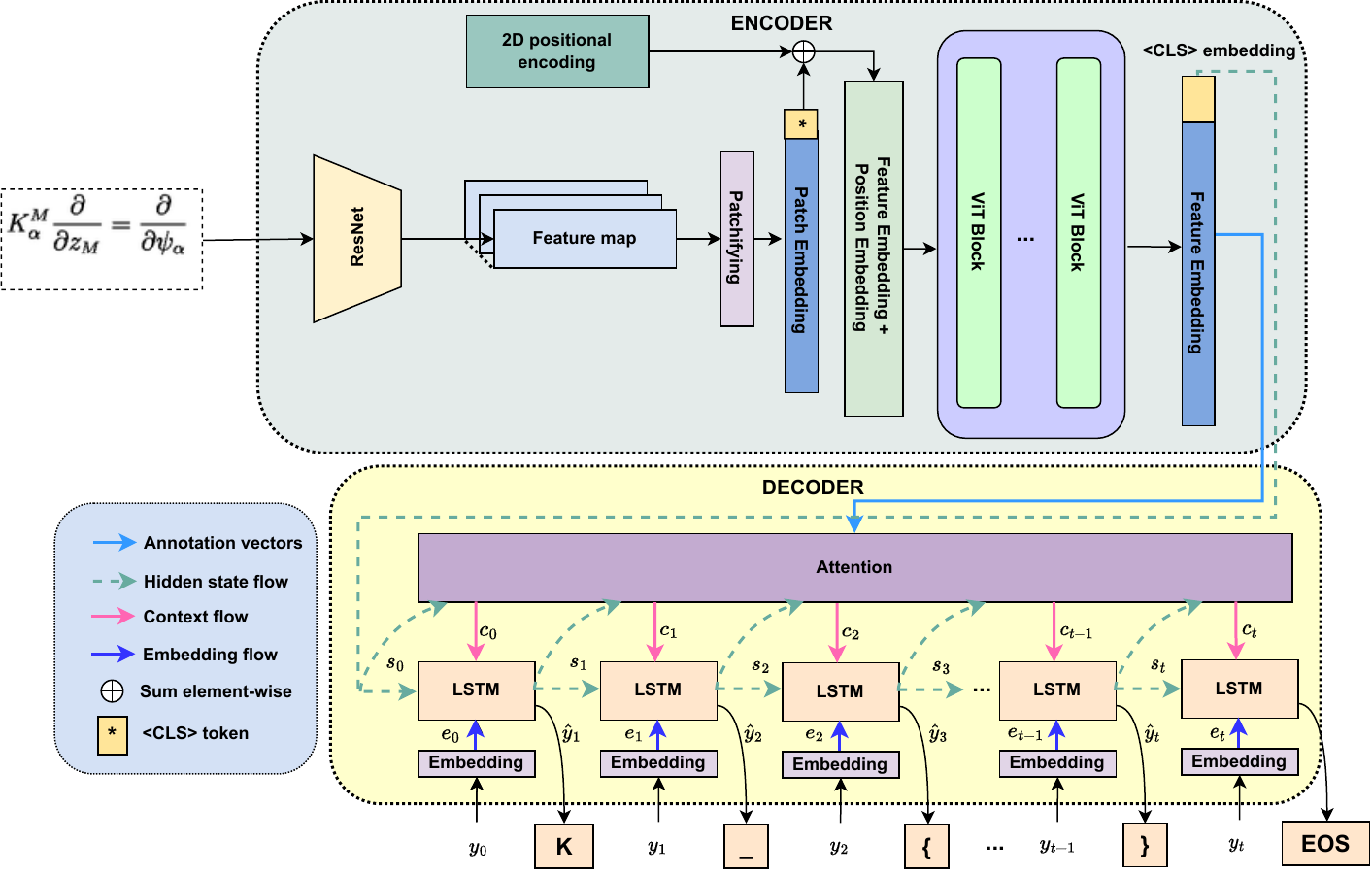}
     \hspace{4em}
    \includegraphics[width=0.12\textwidth]{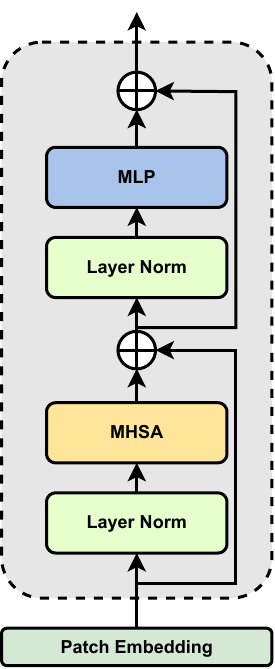}
     \\
    \makebox[0.6\textwidth]{\small (a) The pipeline of proposed architecture}
    \hspace{4em}
    \makebox[0.12\textwidth]{\small (b) An illustration of ViT block}
    \caption{The proposed architecture pipeline and illustration of ViT block.}
    \label{fig:detail}
\end{figure*}

Mathematical expression recognition can be considered as a sequence prediction problem where the input is a grayscale image $\mathbf{X} \in \mathbb{R} ^ {H\times W}$ and the LaTeX ground truth sequence $\mathbf{Y}=\left\{y_{1}, y_{2}, \cdots, y_{\tau}\right\}$ with vocabulary of size $K$ and the length of the sequence is $\tau$. The goal of our model is to convert the input image into the corresponding LaTeX sequence by finding a mapping function $f$ such that $f(\mathbf{X}) = \mathbf{Y}$. In practice, we can just find a function $f^{\prime}$ which is an approximation of function $f$. In this paper, we achieve this purpose by training our model using a dataset of pairs of the image-LaTeX sequences in a supervised manner. Fig. \ref{fig:detail}a demonstrates our proposed architecture including HVT for the encoder and coverage attention for the decoder.

\subsection{Hybrid Vision Transformer as encoder}
Our HVT consists of two modules: First, a CNN is considered as a backbone to extract high-level feature from input image. Second, a context modeling module consists of many ViT block stacks together to further enhance feature embedding by modeling global information and capture long-range dependencies between different feature of the feature maps.

The ViT can perform feature representation of image data for image classification task as shown in \cite{dosovitskiy2020image} by their capability of learning internal relationship between pixels in the images using self-attention mechanism without the need of stacking multiple CNN layers. Therefore, ViT can be served as a perfect encoder in the encoder-decoder framework. While recent works are still based on a recurrent neural network to model context information from the feature maps, one of the most commonly used is bidirectional LSTM (BiLSTM) which can combine the context in both directions of the feature maps. However, BiLSTM will become a bottleneck of the whole architecture due to the sequential design.

Due to the missing of inductive bias in localization compared to CNN as claimed in \cite{dosovitskiy2020image}, ViT needs a lot of training data to attend on small distances like CNN \cite{raghu2021vision}. To make our model converge easily, we add a supporting CNN backbone before the ViT blocks to encode the image's local regions into high-level features.
\subsubsection{Backbone}
We design our ResNet model based on \cite{cheng2017focusing}. Our ResNet-based backbone consists of 32 layers with 4 ResNet blocks. In order to handle text-based images appropriately, stride values at third pooling layer and sixth convolutional layer are changed to (1,2) instead of (2,2) so that feature maps can have a larger width making them easier to cover a correct receptive field of a symbol.

Specifically, give an image $\mathbf{X} \in \mathbb{R}^{H \times W}$, the corresponding output feature maps is $\mathcal{F} \in \mathbb{R} ^ {H_{f} \times W_{f} \times C}$, where $H_{f} = \frac{H}{32} + 1$, $W_{f} = \frac{W}{4} + 1$, $C$ is number of output channels.
\subsubsection{Vision Transformer}
We follow the standard design of ViT from \cite{dosovitskiy2020image} as shown in Fig. \ref{fig:detail}b, such that each ViT block contains a self-attention layer to calculate attention probabilities between \textit{query} vector $\mathbf{q}$ and \textit{key} matrix $\mathbf{K}$ and a feed-forward-network (FFN) which consists of two multilayer perceptron layers (MLP). In order to pay attention to different subspace of different symbols' positions, we further apply multi-head self-attention instead of single-head. Similar to the original Transformer \cite{vaswani2017attention}, ViT uses LayerNorm (LN) \cite{ba2016layer} to stabilize the learning process. Specifically, given $\mathcal{F} \in \mathbb{R} ^ {H_{f} \times W_{f} \times C}$ as the output feature map from the backbone, in order to convert $\mathcal{F}$ to the correspond representation of input of ViT which is a sequence of 1D token embeddings, we further apply a CNN layer with kernel size of $p \times p$ and strides of $p \times p$, where $p$ is defined as the patch size of a patch image, to create a new feature map $\overline{\mathcal{F}} \in \mathbb{R}^{\frac{H_f}{p}\times\frac{W_f}{p}}\times D$, where $D$ is embedding dimensions. The result is finally flattened into a sequence of tokens $\mathbf{E} \in \mathbb{R}^{N \times D}$ known as patch embeddings with $N = \frac{H_f\times W_f}{p^2}$ is the number of patches.\\
ViT also provides an additional token $\mathbf{x}_{class}$ as a learnable embedding which is called [CLS] token inspired by the idea used in BERT \cite{devlin2018bert}, [CLS] token is concatenated with other spatial tokens in the current patch embeddings $\mathbf{E}$ , in the training process, information is forced to flow from all other tokens to [CLS] token through a self-attention mechanism, thus [CLS] token embedding can be considered as a global representation of image features and can be used as an initial hidden state for model's decoder instead of using whole encoder's output feature maps. Moreover, due to the permutation-invariant of self-attention that treats all tokens in sequence as bag-of-word, positional embeddings $\mathbf{E_{p o s}}$ with the same dimension D as patch embeddings are incorporated with patch embeddings to provide position information. The final vector $\mathbf{H_0}$ calculated by Eqs. \ref{eq:vit} is considered as the input vector to ViT.
\begin{align}\label{eq:vit}
    \mathbf{H}_{0} =\left[\mathbf{x}_{\text{class}}, \mathbf{E}\right]+\mathbf{E}_{\text{pos}}
\end{align}
\noindent
Where $\mathbf{E} \in \mathbb{R}^{N \times D}, \mathbf{E}_{p o s} \in \mathbb{R}^{(N+1) \times D}$. Given L ViT blocks of our HVT, Eq. \ref{eq: vitalgo} performs two general steps in the total process of a single ViT block. The MHSA layer first processes the input sequence to mix the information in all tokens in a global context manner. Therefore, one token can accumulate information about other tokens' spatial or semantic features throughout the training process. In mathematical expression recognition, MHSA helps model learn to extract spatial structure for structure analysis and semantic information for symbol recognition. The second step of this process is to go through an FFN. An FFN combines two MLP layers, one layer transforms hidden embeddings from D to 4D dimension and one layer converts 4D back to D dimension. The FFN helps integrate information independently in each token of the sequence.

Concretely, given input sequence $\mathbf{H}_{0} \in \mathbb{R}^{(N+1)\times D}$ we first project it into query $\mathbf{Q}^{h}$, key $\mathbf{K}^{h}$, value $\mathbf{V}^h$ matrix for each head $h \in \left[1, N_{head}\right]$ of MHSA layer using learnable matrix $W_Q^{h} \in \mathbb{R}^{d^{q}\times D}$, $ W_K^{h} \in \mathbb{R}^{d^{k}\times D} $, $W_V^{h} \in \mathbb{R}^{d^{v}\times D}$, in our case $d^q = d^k = d^v = \frac{D}{N_{head}}$. An MHSA layer, as shown in Eq. \ref{eq: sa5} performs self-attention on multiple heads and concatenates the result of all heads together, then projects back to the D dimension using $W_O \in \mathbb{R}^{N_{head}.d^{v} \times D}$. The final output in Eq. \ref{eq: sa4} consists of N + 1 elements, including N spatial embedding vectors of the image $\{h^1_L, h^2_L, \cdots, h^{N}_{L}\}$ called annotation vectors $\mathbf{A}$ and [CLS] token embedding (i.e. $h^{0}_{L}$).

\begin{align}
\left\{\begin{array}{l}
 \mathbf{H}_{\ell}^{\prime} =\operatorname{MHSA}\left(\operatorname{LN}\left(\mathbf{H}_{\ell-1}\right)\right)+\mathbf{H}_{\ell-1} \\
\mathbf{H}_{\ell} =\operatorname{FFN}\left(\operatorname{LN}\left(\mathbf{H}_{\ell}^{\prime}\right)\right)+\mathbf{H}_{\ell}^{\prime}
\end{array} \quad \ell \in \left[1, L\right] \right.
\label{eq: vitalgo}
\end{align}

\begin{align}
\operatorname{MHSA_{\ell}(Q, K, V)} & = \left[SA^{1}_{\ell}, SA^{2}_{\ell}, \cdots ,SA^{H}_{\ell}\right]\times W_{O}
\label{eq: sa5}
\end{align}
\begin{align}
\left[h^{0}_{L}, h^{1}_{L}, h^{2}_{L}, \cdots, h^{N}_{L}\right] & = \mathbf{H}_{L}
\label{eq: sa4}
\end{align}


\subsubsection{2D Positional Encoding}
Unlike the natural image, a math image has a strong semantic correlation between different components in the formula, such position information about nested or hierarchical components needs to be reserved carefully. Under this assumption, we propose 2DPE which using 2D sinusoidal positional encoding similar to \cite{chen2021empirical}. 2DPE performs 1D positional encoding from \cite{vaswani2017attention} on two dimensions of the input features and then concatenates them to gain the final output.\\
Concretely, $\left(H, W\right)$ is height and width of the feature maps and $D$ is embedding dimension, the position embedding of the feature maps is calculated by our 2DPE using Eq. \ref{eq:pos}:
\begin{align}
    \label{eq:pos}
    \operatorname{2DPE}(H, W, D) = \left[\operatorname{PE}(h, i), \operatorname{PE}(w, j)\right]\
\end{align}
\noindent
Where $i \in \left[0,D/2\right), j \in \left[D/2, D\right)$, $\left[.\right]$ denotes concatenate operation. The embedding for each dimension can be obtained as follow: 
\begin{align}
    \label{eq:1dpos}
    \left\{\begin{array}{l}
    \operatorname{PE}{(pos, t)} =\sin \left(\operatorname{pos} / 10000^{t / \frac{D}{4}}\right) \\
\operatorname{PE}{(pos, t + \frac{D}{4})} =\cos \left(\operatorname{pos} / 10000^{t / \frac{D}{4}}\right)
    \end{array}\right.
\end{align}
\noindent
Where $t \in \left[0,D/4\right)$, $\operatorname{PE}$ denotes 1D sinusoidal positional encoding, and $\operatorname{pos}$ denotes any position in the horizontal (w) or vertical direction (h).
\subsection{Coverage attention for Decoder}
To deal with variable input length of math image and variable output length of LaTeX sequence, we choose RNN-based with attention mechanism \cite{shi2018aster} as our decoder. At every time step $t$, our decoder calculate a fixed-size intermediate vector also known as context vector $\mathbf{c}_t \in \mathbb{R}^D$ based on a weighted sum between the annotation vectors $\mathbf{A}$ and the attention weight $\pmb{\alpha}_t \in \mathbb{R}^N$. By using $\mathbf{c}_t$, our decoder can generate a suitable LaTeX symbol and also be independent of variable input-output length. We consider using unidirectional LSTM instead of traditional recurrent neural networks to overcome the vanishing gradient problem.\\
At training phase, each groundtruth (GT) token $\mathbf{y}$ is first transformed using an embedding layer as shown in Fig. \ref{fig:detail} into a vector $\mathbf{e}$ so that tokens which have high correlation (e.g. '[', ']') can have similar embedding.\\
At time step 0, we choose vector embedding of [CLS] token (i.e. $h^{0}_L$) as the initial hidden state. Different from \cite{tao2021trig} approach, we further apply an MLP layer to $h^{0}_L$ to integrate more information. Coverage attention similar to \cite{zhang2017watch} is applied to our decoder to make it more robust with under-parsing and over-parsing problems. Coverage vector is created by summing all previous attention weights $\pmb{\alpha}$ and performing a convolution operation (Conv) to aggregate information of the history alignments as shown in Eq. \ref{eq: coverage} and therefore it can guide the decoder on future prediction to put attention on appropriate regions. 
\begin{align}
\label{eq: coverage}
    \left\{\begin{array}{l}
    \pmb{\beta}_{t} = \sum_{l=1}^{t-1}\pmb{\alpha}_{l} \\
    \mathbf{f}_{t} = \operatorname{Conv}(\pmb{\beta}_{t})
    \end{array}\right.
\end{align}
\noindent
Where $\mathbf{f}_{t}$ indicates coverage vector at step t, $\mathbf{\beta}_{0}$ is set to vector 0. 
In summary, the output $\mathbf{\hat{y}}_{t} \in \mathbb{R}^{K}$ at step t is calculated by Eq. \ref{eq: decoder}. 
\begin{align}
    \label{eq: decoder}
    \left\{\begin{array}{l}
    \mathbf{c_{t}} = \operatorname{Attention}(\mathbf{s}_{t-1} , \mathbf{A}, \mathbf{f}_{t})) \\ 
    \mathbf{s}_{t} = \operatorname{RNN}(\mathbf{s}_{t-1}, \left[\mathbf{c}_{t}, \mathbf{e}_{t}\right]) \\
    \mathbf{\hat{y}_{t}} = \operatorname{Softmax}(\mathbf{s}_{t})
    \end{array}\right.
\end{align}
\noindent
Where $\operatorname{Attention}$ is attention operation as described in \cite{zhang2017watch}, $\operatorname{RNN}$ is a recurrent layer specifically a LSTM layer, $\operatorname{Softmax}$ is a softmax layer to output probability distribution, $\left[.\right]$ denotes concatenate operation, $\mathbf{s}_{j}$ is decoder hidden state of step $j \in \left[0, \tau\right]$, $\mathbf{e}_{t}$ is embedding vector of $\mathbf{y}_{t}$ GT token. At time step 0, $\mathbf{s}_{0} = \operatorname{MLP}(\mathbf{h}^{0}_{L}$).
\section{Experiments}
This section presents our experiment on the benchmark dataset and compares it with other SOTA models. We also perform an extensive ablation study to analyze the effectiveness of each component in our model, and finally, some visualization of the attention weight in both encoder and decoder.   
\subsection{Dataset}
We choose to use a benchmark dataset IM2LATEX-100k \cite{deng2017image} created by crawling from 60000 research papers on arXiv. The dataset contains 103,556 LaTeX representation of mathematical expressions in total. The length of each LaTeX sequence is change from 38 to 997. Each sequence is rendered to PDF format using \textit{pdflatex} compiler and converted to PNG image in grayscale format using \textit{ImageMagick}. The final dataset is split into 3 partitions including train set with 83,883 formulas, validation set with 9319 formulas, and test set with 10,354 formulas.\\
We follow the same preprocessing strategy as \cite{deng2016you} by applying a parsing algorithm on raw LaTeX sources to create tokenized LaTeX labels. We then create a vocabulary $\mathrm {V}$ from all unique LaTeX tokens with three addition tokens, including [SOS], [EOS], and [PAD] tokens. Our vocabulary $\mathrm{V}$ contains 499 tokens. All images of the same size will be grouped into the same bucket. This way can help to reserve the 2D structure of math images, which is different from normal text images.
\subsection{Implementation details}
\subsubsection{Architecture}
For our HVT configuration, we set the number of output channels of ResNet-based backbone to $C = 512$. Different from the original setting of ViT from \cite{dosovitskiy2020image}, we reduce the number of heads to $N_{head} = 8$ and we only use $L = 6$ ViT blocks to encode the image's feature, we choose the dimension the patch embedding to $D = 256$, dimension of FFN layer to $d_{ffn} = 1024$. ince we perform patchify on feature maps, we choose a small patch size $p=2$. For the decoder, we apply a filter with a kernel size of $5\times 5$ and an output channel of $128$ to the coverage vector; both the hidden state of LSTM and the embedding size of the input token are set to $d_{emb} = 256$. We adopt a dropout \cite{srivastava2014dropout} technique with drop rate 0.1 as a regularization method to reduce over-fitting.
\subsubsection{Training and inference}
In the training phase, at every time step, the decoder will receive the embedding of the ground truth token, also known as teacher forcing. In the inference phase, the output LaTeX sequence is predicted token by token at every time step. Moreover, to prevent the predicted output from sampling from a sub-optimal distribution, we use beam search with beam size set to $5$ to get the output token. The entire model is trained from scratch without using any pretrained weight for 300K iterations with a batch size of $32$ using AdamW \cite{loshchilov2017decoupled} optimizer with an initial learning rate set to $5\times 10^{-4}$ and the decay rate of $2\times10^{-6}$. After every step, the learning rate is adjusted by using a warm-up cosine schedule. A simple data augmentation strategy that includes random scale and rotation is adopted to better optimize ViT. All experiments are implemented using PyTorch and conducted on GPU NVIDIA V100 32GB.
\subsubsection{Evaluation}
We also consider evaluating our model's performance using text- and image-based metrics. For text-based metrics, we use BLEU-4 score \cite{papineni2002bleu}, text edit distance (TED) which compute the Levenshtein distance between the GT sequence and the prediction at token level, and the sequence accuracy (Acc) which return 1 if the prediction and GT are exactly the same else 0. For image-based metrics, we evaluate on the rendered images of predicted LaTeX sequences and GT images using the same metrics used in \cite{deng2016you} including image edit distance (IED), and exact match accuracy without spaces (EMA w/o space). 

\subsection{Compare between other SOTA methods}
We demonstrate the effectiveness of our proposed model by comparing it with previous methods on IM2LATEX-100K test set. Our method achieves a better result than \cite{pang2021global} which proves the effectiveness of using ViT compared to a global context module from \cite{pang2021global}. Especially, for image-base metric as EMA, our method obtain a significant improvement of about $2.4\%$ compared to \cite{zhang2019improved} which proves the potential of our method in capturing the image's structure. Besides, Acc result which is very low compare to other metrics has shown the ambiguities in LaTeX grammar where many LaTeX sequences can represent the same visual structure.
	\begin{table}
		\centering
		\caption{Comparison between different methods on IM2LATEX-100K.}
		\label{tab:sota}       
	\begin{threeparttable}
		\scalebox{0.9}{
		\centering\noindent
	\renewcommand{\arraystretch}{1.2}
   	 \begin{adjustbox}{width=0.5\textwidth}
			\begin{tabular}{lcccc}
				\hline\noalign{\smallskip}
				\multirow{2}{*}{Method} & \multirow{2}{*}{Acc} & \multirow{2}{*}{BLEU-4}  & \multirow{2}{*}{EDA} & \multirow{2}{*}{EMA} \cr
				&&&& (w/o space) \cr 
				\noalign{\smallskip}\hline\noalign{\smallskip}
                Global Context \cite{pang2021global} & - & 89.72 & 90.07 & 82.13 \\
				Double Attention \cite{zhang2019improved} & - & 89.4 & 90.9 & 84.1 \\
				MI2LATEX w/o reinforce \cite{wang2021translating}\tnote{*} & - & 89.08 & 91.09 & 82.13 \\
				\rowcolor{gray!20}
				\Frst{Ours} & \Frst{48.39} & \Frst{89.94} & \Frst{92.23} & \Frst{86.48} \\
				\noalign{\smallskip}\hline
		\end{tabular}
	\end{adjustbox}}
	\begin{tablenotes}
   \item[*] \small{We only consider the MI2LATEX version without second \\training phase.}
  \end{tablenotes}
  \end{threeparttable}
	\end{table}

\subsection{Ablation Study}
\subsubsection{Contribution of main components}
To better understand the effectiveness of each component in our proposed method, we extensively compare the performance of the baseline model with other model's versions when we alternately replace each component of the baseline with our proposed ones.\\
For the baseline, we choose VGG \cite{simonyan2014very} architecture as the backbone and do not use any context modeling module, attention mechanism (Attn) without coverage vector similar to \cite{shi2018aster} is used as the prediction head. For feature extraction, we compare VGG backbone with our ResNet-based. For context modeling, we choose BiLSTM from \cite{shi2018aster}, we also experiment on a ViT version which receives 1D feature maps called ViT-1D by collapsing the height of input feature maps into one together with our's ViT-2D to perform the comparison. For prediction, we consider using a transformer decoder from \cite{vaswani2017attention} and our coverage attention (Coverage-Attn) against the baseline. Detail of the all experiment setting and the results of ablation experiments are shown in Table \ref{tab:setting} and \ref{tab:ablation_study}. \\
According to the results, using our ResNet-based instead of VGG as the backbone has improved the baseline by a large margin on all evaluation metrics, especially the accuracy has increased by $15\%$. This suggests that a ResNet model for text recognition is a good choice for our backbone. In the context modeling component, we can observe that adding the ViT-2D module gives the best performance. It shows the ability to model a better long-range dependencies compared to BiLSTM and reserve a 2D spatial structure compared to ViT-1D. Using coverage attention for the baseline instead of \cite{shi2018aster} also gives a better result than using the transformer decoder, which indicates the importance of the coverage vector in keeping the alignment history. 
\begin{table}[thb]
    \centering
    \caption{Detail of ablation experiment setting, where `None' indicate empty module.}
    \noindent
	\renewcommand{\arraystretch}{1.2}
  	 \begin{adjustbox}{width=0.4\textwidth}
     \begin{threeparttable}
       \begin{tabular}{|l|c|c|c|}
        \Xhline{4\arrayrulewidth}
        \multirow{2}{*}{Experiment} & \multicolumn{2}{c|}{Encoder} & \multirow{2}{*}{Pred.} \cr\cline{2-3}
        & Feat. & Context. &
    \cr\Xhline{4\arrayrulewidth}
    Baseline & VGG & None & Attn \\
    V1 & ResNet & None & Attn \\
    V2 & VGG & ViT-1D & Attn \\
    V3 & VGG & BiLSTM & Attn \\
    V4 & VGG & ViT-2D & Attn \\
    V5 & VGG & None & Transformer Decoder \\
    V6 & VGG & None & Coverage-Attn \\
    V7 & ResNet & ViT-2D & Coverage-Attn\\
    \hline
  \end{tabular}
    \end{threeparttable}
    \end{adjustbox}
    \label{tab:setting}
\end{table}
\begin{table}[thb]\centering
    \caption{Ablation experiments on different key components of our approach evaluate on IM2LATEX-100K validation set. Particularly, Feat. denote feature extraction, Context. denote context modeling, and Pred. denote prediction.}
    \label{tab:ablation_study}
    \resizebox{0.4\textwidth}{!}{
    \Large
    \begin{tabular}{*{2}{l}||*{1}{c}||*{1}{c}||*{1}{c}||*{1}{c}}
        \toprule
        \multicolumn{1}{c}{} & \multicolumn{1}{c||}{} & Acc & BLEU-4 & TED & params\\
        Component & Experiment  & $\%$ & $\%$ & $\%$ & $\times 10^{6}$\\
        \midrule
        \multirow{2}{*}{{Feat.}}  & Baseline  & 22.15 & 79.12 & 77.32 & 7.05 \cr %
         & \Scnd{V1} & \Scnd{37.55} & \Scnd{85.90} & \Scnd{89.90} & \Scnd{45.76} \cr
        \midrule
        \multirow{3}{*}{{Context.}} & Baseline  & 22.15 & 79.12 & 77.32 & 7.05 \cr
        & V2 & 34.97 & 87.73 & 92.23 & 12.13 \cr
         & V3 & 43.69 & 90.61 & 94.52 & 9.75\cr %
         & \Scnd{V4} & \Scnd{44.42} & \Scnd{91.48} & \Scnd{94.88} & \Scnd{12.19} \cr %
        \midrule
        \multirow{2}{*}{{Pred.}} & Baseline  & 22.15 & 79.12 & 77.32 & 7.05 \cr
        & V5 & 42.06 & 90.12 & 93.04 & \Scnd{15.52} \cr %
         & \Scnd{V6} & \Scnd{42.58} & \Scnd{90.23} & \Scnd{93.12} & 7.09 \cr %
        \midrule
        \rowcolor{gray!20}
        \Frst{Overall} & \Frst{V7} & \Frst{49.30} & \Frst{92.33} & \Frst{95.60} & \Frst{50.95} \\ %
        \bottomrule
    \end{tabular}}
\end{table}

\subsubsection{Contribution of 2d positional encoding}
To study the impact of 2D positional encoding, we compare the result when using 1D positional encoding similar to \cite{dosovitskiy2020image, vaswani2017attention} and our 2DPE before entering the ViT. Table \ref{tab:pos_effect} shows that using our 2DPE has a better result on all metrics. Significantly, the EMA has improved by approximately $4\%$. Experiment's result is evaluated on IM2LATEX-100K test set.

\begin{table}[thb]\centering
    \caption{Comparison between two positional encoding method when apply to our proposed method.}
    \label{tab:pos_effect}
    \resizebox{0.4\textwidth}{!}{
    \large
    \begin{tabular}{|c|cccc|}
        \toprule
        Positional Encoding &  Acc &  BLEU-4 &  IED & EMA (w/o space) \\
        \midrule
        1D & 45.78 & 88.84 & 89.31 & 82.93 \\
        2D & \Frst{48.39} & \Frst{89.94} & \Frst{92.23} & \Frst{86.48} \\
        \bottomrule
    \end{tabular}
    }
\end{table}
\subsubsection{Contribution of [CLS] token embedding vector}
To study the impact of using [CLS] token embedding as an initial hidden state for the decoder, we compare the result when not using initial embedding and when using it. Table \ref{tab:init_effect} shows that 2D positional encoding has obtained a better result than the 1D positional encoding approach. Experiment's result is evaluated on IM2LATEX-100K test set.
\begin{table}[thb]\centering
    \caption{Comparison on two initial hidden state settings when apply to our proposed method.}
    \label{tab:init_effect}
    \resizebox{0.4\textwidth}{!}{
    \large
    \begin{tabular}{|c|cccc|}
        \toprule
        Use init &  Acc &  BLEU-4 &  IED & EMA (w/o space) \\
        \midrule
        \cmark & \Frst{48.39} & \Frst{89.94} & \Frst{92.23} & \Frst{86.48} \\
        \xmark & 43.87 & 81.73 & 89.36 & 81.02 \\
        \bottomrule
    \end{tabular}
    }
\end{table}

\subsection{Discussion}
\subsubsection{The effect of LaTeX sequence length}
\begin{figure}[thb] \centering
    \includegraphics[width=0.5\textwidth]{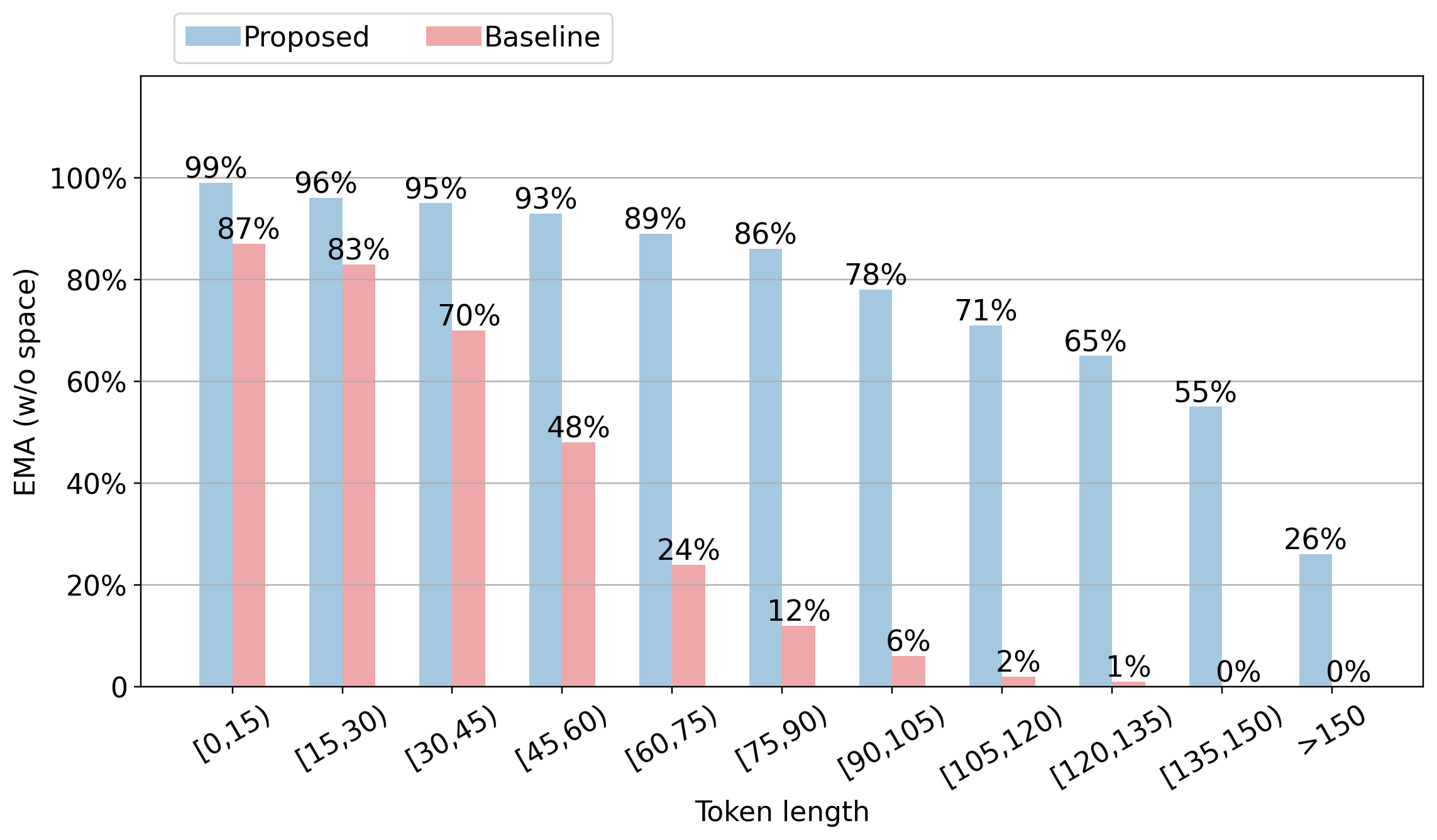}
    \caption{Comparison between our proposed method and the baseline model at different sequence length.} \label{fig:seqlen}
\end{figure}

Given the ground truth of LaTeX sequences, we manually group them into different groups based on their length to investigate the effect of LaTeX sequence length on the performance of our method. To prove the consistency of our method to the sequence length, we compare the average EMA (w/o space) between our method and the baseline model on different group lengths. Fig. \ref{fig:seqlen} shows that our method is very robust to the length of LaTeX sequences while the baseline's performance decreases significantly. Our method still has the EMA of more than $71\%$ when sequence length is more significant than $100$ and has $26\%$ for a sequence with more than $150$ tokens.


\subsubsection{Encoder visualization}
\begin{figure}[thb]
    \centering
    \includegraphics[keepaspectratio,width=0.5\textwidth]{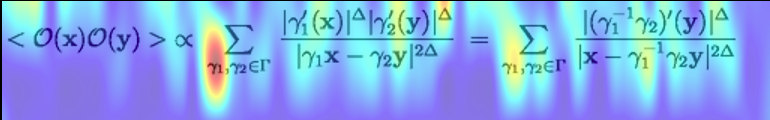}
    \caption{Examples of self-attention map of [CLS] token embedding.} \label{fig:encoder_attn}
\end{figure}

The input of ViT is the sequence of embedding vector of spatial locations in the input image plus the embedding of [CLS] token. Fig. \ref{fig:encoder_attn} shows the self-attention map of the [CLS] token embedding when attending to all other spatial embedding vectors. This visualization has confirmed the usefulness of using our HVT in modeling global information between different math symbols in the images.

\subsubsection{Decoder visualization}
\begin{figure}[thb]
    \centering
    \includegraphics[width=\linewidth]{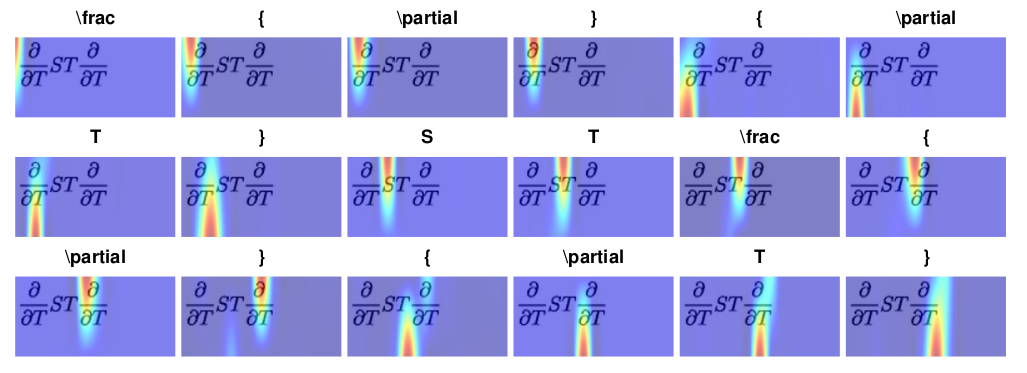}
    \caption{Visualization of a step-by-step decoding process using our method on example math expression image.}
    \label{fig:attention visulation}
\end{figure}

We visualize the step-by-step decoding process using coverage attention on an example math expressions of IM2LATEX-100K testing in Fig. \ref{fig:attention visulation}. At each step, the attention map shows that the model correctly aligns some local region on the image to the corresponding math symbol.

\subsubsection{Limitation}
\begin{figure}[thb]
\centering
\includegraphics[width=0.5\textwidth]{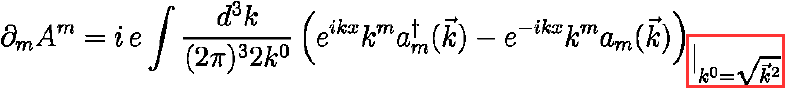}
\\
\makebox[0.5\textwidth]{\small (a) Groundtruth}
\includegraphics[width=0.5\textwidth]{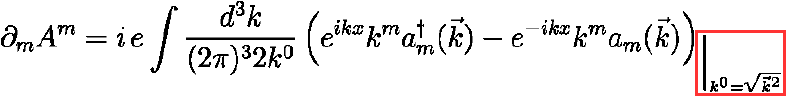}
\\
\makebox[0.5\textwidth]{\small (b) Prediction}
\caption{An example of our model's prediction about the correctness in spatial structure and symbol correlation but misunderstanding in the syntactic relationships.}
\label{fig:limit}
\end{figure}

Fig. \ref{fig:limit} has shown that despite the strong ability to capture the global dependencies and correlation between symbols in math image, and the capable symbol recognition mechanism through coverage attention help the model to implicitly learn about the grammar rules, it still suffers from the lack of specific knowledge about grammar.
\section{Conclusion and Future work}
In this paper, we have proposed a novel Hybrid Vision Transformer approach combined with the embedding vector of [CLS] token as the initial hidden state of the decoder allowing the model to extract more sophisticated relationships. Our architecture includes three main stages, which are feature extraction, context modeling and prediction.
Our approach has proved the effectiveness when compared to other approaches. Our model has achieved a SOTA performance on the well-know public dataset IM2LATEX-100K. In the future, our research will focus on appending synthetic LaTeX information into the Seq2seq model to better handle more complicated math expression structure. Besides, we will build a complete system to be able to provide products to users.
\section*{Acknowledgment}
\indent We acknowledge Ho Chi Minh City University of Technology (HCMUT), VNU-HCM for supporting this study.

\bibliographystyle{IEEEtran}

\bibliography{egbib}

\end{document}